\documentclass[10pt,twocolumn,letterpaper]{article}

\usepackage[pagenumbers]{cvpr}

\usepackage{graphicx}
\usepackage{amsmath}
\usepackage{amssymb}
\usepackage{bm}
\usepackage{booktabs}
\usepackage{multicol}
\usepackage{multirow}
\usepackage{amssymb}
\usepackage{soul}
\usepackage{appendix}
\usepackage{arydshln}
\usepackage{colortbl}

\newcommand{\minisection}[1]{\vspace{1mm}\noindent{\textbf{#1}~}}

\usepackage[pagebackref,breaklinks,colorlinks]{hyperref}

\usepackage[capitalize]{cleveref}
\crefname{section}{Sec.}{Secs.}
\Crefname{section}{Section}{Sections}
\Crefname{table}{Table}{Tables}
\crefname{table}{Tab.}{Tabs.}
\begin{document}

\title{Sylph: A Hypernetwork Framework\\ for Incremental Few-shot Object Detection}

\author{Li Yin \quad Juan M Perez-Rua \quad Kevin J Liang  \\
Meta AI  \\
{\tt\footnotesize \{liyin, jmpr, kevinjliang\} @fb.com}}
\maketitle
\graphicspath{ {./fig/} }

\begin{abstract}
We study the challenging incremental few-shot object detection (iFSD) setting. Recently, hypernetwork-based approaches have been studied in the context of continuous and finetune-free iFSD with limited success. 
We take a closer look at important design choices of such methods, leading to several key improvements and resulting in a more accurate and flexible framework, which we call \textit{Sylph}.
In particular, we demonstrate the effectiveness of decoupling object classification from localization by leveraging a base detector that is pretrained for class-agnostic localization on a large-scale dataset. 
Contrary to what previous results have suggested, we show that with a carefully designed class-conditional hypernetwork, finetune-free iFSD can be highly effective, especially when a large number of base categories with abundant data are available for meta-training, almost approaching alternatives that undergo test-time-training. 
This result is even more significant considering its many practical advantages: (1) incrementally learning new classes in sequence without additional training, (2) detecting both novel and seen classes in a single pass, and (3) no forgetting of previously seen classes. We benchmark our model on both COCO and LVIS, reporting as high as $17\%$ AP on the long-tail rare classes on LVIS, indicating the promise of hypernetwork-based iFSD.
\end{abstract}

%%%%%%%%% BODY TEXT
\vspace{-2mm}
\section{Introduction}
\vspace{-2mm}
While advances in deep learning have led to significant progress in computer vision~\cite{krizhevsky2012imagenet, girshick2014rich, he2016deep, he2017mask}, much of this success has relied upon large-scale data collection and annotation~\cite{deng2009imagenet, lin2014microsoft, gupta2019lvis, grauman2022ego4d}, a process that is both labor-intensive and time-consuming, and does not scale well with the number of categories. This is especially true for object detection~\cite{girshick2014rich, he2017mask, liu2020deep}, particularly for the long tail of object categories, where data may be scarcer~\cite{gupta2019lvis}. 
As a result, few-shot learning of object detectors (FSD)~\cite{kang2019few, yan2019meta, wang2020frustratingly, zhang2021meta} has become a recent topic of interest. 

While learning a novel class from only a few samples alone is a challenging problem, the task can be made simpler by leveraging known classes with abundant data (commonly referred to as \textit{base} classes), whose structure can be used as a prior for knowledge transfer.
The few previous FSD works have approached this primarily in two ways.
The first is fine-tuning~\cite{wang2020frustratingly, qiao2021defrcn}, where a model is first pretrained on the base classes and then fine-tuned on a small balanced set of data from both the base and novel classes, a form of test-time training~\cite{sun2020test}.
Although simple, it has difficulty scaling to many real-world applications due to its computational and memory requirements.
An alternate strategy is taking a meta-learning approach~\cite{zhang2021meta}.
Meta-learning approaches frame the problem as ``learning to learn''~\cite{xiao2020few, zhang2021meta, thrun2012learning, chen2021dual, fan2020few, perez2020incremental, li2021beyond}, training the model episodically to induce fast adaptation to novel classes.

However, many FSD methods focus on the limited set-up where only novel categories are to be detected. These methods often fail to preserve the original detector performance on base categories~\cite{fan2020few, zhang2021meta, chen2021dual, li2021beyond} or forget about the ones it was initially trained on~\cite{wang2020frustratingly}.
Given the ever-changing nature of the real-world, a desirable property of machine learning systems is the ability to incrementally learn new concepts without revisiting previous ones and not forgetting them~\cite{parisi2019continual, mehta2021continual}. 
Humans are able to achieve such feat, learning novel concepts not only without forgetting but reusing such knowledge~\cite{pinker2003mind}. 
Conventional supervised learning struggles with incrementally presented data, tending to suffer catastrophic forgetting~\cite{McCloskey1989, Ratcliff1990}.
An alternative is studying all the available data every time new concepts arrive, commonly referred to as ``joint training''~\cite{gupta2019lvis}, but such a paradigm imposes a slow development cycle, requiring significant data collection efforts for the new concepts and expensive large-scale training (and re-training).

Instead, we seek an object detection model capable of learning new classes from a few shots in a fast, scalable manner without forgetting previously seen classes, a setting commonly referred to as incremental few-shot detection (iFSD).
ONCE~\cite{perez2020incremental}, a meta-learning approach to FSD, is of particular interest due to its hypernetwork-based class-conditional design. 
ONCE is able to enroll novel categories without affecting its ability to remember base classes. 
We use a base detector and hypernetwork architecture similar to ONCE, but with a few key design differences: (1)
ONCE, along several other recent works~\cite{kang2019few, wang2020frustratingly, zhang2021meta}, attempts to directly produce (via training or hypernet) the parameters of a localization regression model that transforms the query sample feature maps into the output bounding boxes, all from the few available training samples. We find this to be unnecessary and potentially harmful, as the task can be significantly simplified by decoupling localization from classification.  
To achieve this goal, we leverage a base detector with class-agnostic localization capability pretrained on abundant base class data. 
(2) We study the class-conditional hypernetwork's behavior, making some key changes to the structure and adding normalization to the predicted parameters, resulting in much higher accuracy.
 
\begin{figure*}[t]
    \centering
    \includegraphics[width=0.85\textwidth]{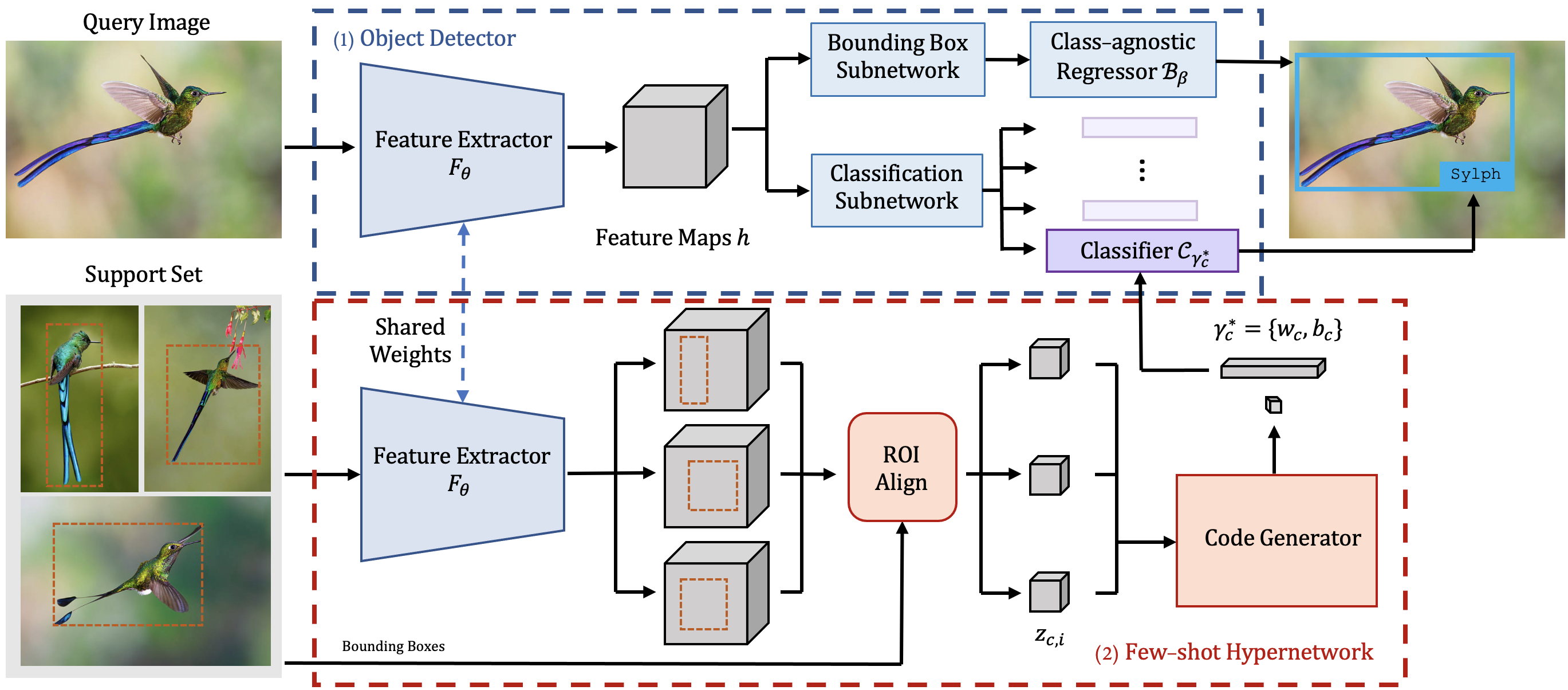}
    \caption{\textbf{The Sylph Framework.} Sylph is composed of a base object detector and a few-shot hypernetwork, whose Code Generator consists of a Code Predictor Head and Code Process Module (detailed in Section.~\ref{few-shot hypernetwork}). The dashed arrow indicates weight sharing.}
    \label{fig:sylph}
\end{figure*}

With an architecture that can swiftly adapt to the long tail of classes from few shots, we name our framework \textit{Sylph}, after the nimble long-tailed hummingbird (Figure~\ref{fig:sylph}).
We present extensive evaluations that empirically demonstrate the benefits of our design, showing that Sylph is more effective than ONCE~\cite{perez2020incremental} (our main baseline) across all the reported datasets and evaluation regimes. On the challenging LVIS few-shot learning benchmark in particular, we show that Sylph is superior by a margin of $8\%$ points.  

\vspace{-2mm}
\section{Related Work}
\vspace{-2mm}
\minisection{Object Detection } Object detection is the task of simultaneously localizing and classifying objects within a scene.
Most modern object detectors consist of a convolutional feature extractor~\cite{krizhevsky2012imagenet, simonyan2014very, he2016deep} followed by various mechanisms or networks to predict classes and some form of bounding box coordinates~\cite{huang2017speed}.
Detectors that first generate region proposals during inference are often referred to as two-stage detectors~\cite{girshick2014rich, girshick2015fast, ren2015faster, dai2016r, he2017mask}, while ones that directly predict class and localization from the convolutional feature maps are considered single-stage detectors~\cite{liu2016ssd, redmon2016you, lin2017focal, duan2019centernet, tian2019fcos}.
Single-stage detectors have the advantage of having simpler implementations and faster inference speeds, and recent advances have increased their accuracy to be competitive with two-stage models~\cite{liu2020deep}, which had previously been the primary advantage of such models.
Throughout this work, we choose to primarily use FCOS~\cite{tian2019fcos} as our base detector due to its strong performance and class-agnostic localization based on ``centerness'' and intersection-over-union (IoU) losses; 
this allows for better generalization and high recall on novel unseen classes~\cite{kim2021learning}, especially when trained on large-scale datasets. 

\minisection{Few-shot Learning } While many supervised learning approaches assume a large number of samples from the data distribution, such methods risk overfitting when the model has only a few samples to learn from. 
Given the costs of collecting, annotating, and training models with large amounts of data, few-shot learning has become an active research direction, with image classification as the most common task.
Many recent approaches take a meta-learning strategy~\cite{vanschoren2018meta}.
Optimization-based approaches produce models that can quickly learn from few samples~\cite{ravi2016optimization, finn2017model, nichol2018first}.
Metric-learning methods learn an embedding function that induces a space where samples can be compared with nearest neighbors or other such simple algorithms~\cite{vinyals2016matching, snell2017prototypical, sung2018learning, liang2022few}. 
Hypernetworks have also been used to predict model parameters for new classes from limited samples~\cite{bertinetto2016learning, garnelo2018conditional, gidaris2018dynamic, qiao2018few, wang2019tafe, rangrej2021revisiting}.
We use a hypernetwork in our model to predict convolutional kernels for novel object classification.
Such a strategy requires zero training during inference time and can easily scale to an arbitrary number of classes.

\minisection{Few-shot Object Detection and Beyond } Most neural networks are trained with stochastic gradient descent, which often assumes the training data are drawn independently and identically distributed (\textit{i.i.d.}). However, this \textit{i.i.d.} assumption is violated in the practical scenario where few-shot categories are seen only after the model have been trained for a set of base categories. In such situations, catastrophic forgetting~\cite{goodfellow2013empirical, farquhar2018towards} can occur: the model suffers severe degradation in performance on the original classes. In image classification, some works have proposed a generalized setting for few-shot learning to tackle this exact situation~\cite{gidaris2018dynamic,qi2018low}. 
Similarly for object detection, recent works have focused on incorporating few-shot categories into a model that has been pretrained with large-scale datasets~\cite{perez2020incremental, wang2020frustratingly, fan2021generalized}. This goes beyond the simpler more traditional few-shot object detection set-up~\cite{qiao2018few, qiao2021defrcn, kang2019few}. More generally, continual object detection~\cite{acharya2020rodeo, perez2020incremental} works attempt to learn to detect new classes through several learning instances without forgetting any of the seen categories.

Of the prior work with the goal of both few-shot and continual learning for object detection, some are continual only in that they do not degrade base class accuracy during a \textit{single} few-shot adaptation to new classes~\cite{fan2021generalized, wang2020frustratingly}.
In contrast, ONCE~\cite{perez2020incremental} considers a setting in which novel classes arrive sequentially and \textit{incrementally}, leading to multiple learning events during which forgetting must be avoided.
We adopt a model architecture that is able to provide such capabilities, as it is more flexible and a better fit for interactively learning new classes from the world.
Methodologically, however, we approach the problem differently, as we 
(1) simplify learning by utilizing a detector with class-agnostic localization rather than trying to learn per-class localization from only a few samples;
(2) leverage a per-class binary classifier to allow incrementally and independently added novel classes to co-exist with previously learned base classes, detecting seen and novel classes in a single pass;
(3) generate both weights and biases for newly added classes, proposing an effective weight normalization to the output of a hypernetwork weight generator that enables stable training and more effective synthesis of class-specific class codes.

\section{Methods}
\vspace{-2mm}
We seek a model that can operate in the incremental few-shot detection (iFSD)~\cite{perez2020incremental} setting: a detector that can flexibly adapt to new classes introduced in sequence from only a few examples, without forgetting any previously seen classes. We differentiate this \textit{continuous} iFSD with \textit{batch} iFSD where novel classes are added in a batch.
Concretely, after being pretrained on a base set of classes $C^b$, the objective is to achieve good performance on a novel class $c^n_t \in C^n$ from a support set of only $K$ shots while maintaining strong performance on $C^b$ and the preceding novel classes $c^n_{t'}~\forall~t'<t$, \textit{without} re-training on data from these previous classes.
As the goal is to learn to adapt to new classes, we assume $C^b \cap C^n = \emptyset$.

\subsection{Sylph}
To achieve the stated objective of iFSD, we introduce Sylph, a framework that can quickly add detection capabilities of new classes, without any additional optimization of model parameters.
Sylph is composed of two primary components (Figure~\ref{fig:sylph}): (1) a base object detector with class-agnostic localization to surface salient objects in an image with high recall and (2) a few-shot hypernetwork to generate class-specific parameters for a per-class binary classifier.
We discuss each of these in detail below.

\subsubsection{Object Detector}
Modern object detection models~\cite{huang2017speed} are often composed of a convolutional backbone $\mathcal{F}_\theta$ followed by a detector head $\mathcal{D}_\phi$.
Given an image $I$, the former produces high-level feature maps $h=\mathcal{F}_\theta (I)$, which can then be used by the detector head to predict both class $c$ and location, as specified by a bounding box $b=(x, y, h, w)$.
Many detection models perform both these tasks in parallel~\cite{girshick2014rich, ren2015faster, liu2016ssd}, predicting the class category and bounding box coordinates from the same features: $\bm  o = \mathcal{D}_\phi(h)$, where $\bm o = [o_1, ..., o_n]$ are predicted objects in $I$, with each object $o_i = [c_i, b_i]$ containing the predicted class label and bounding box.  
We denote the final regression and classification layer as $\mathcal{B_\beta}$ and $\mathcal{C_\gamma}$, which can be a fully connected layer in region-based detection~\cite{ren2015faster} or a convolutional layer in dense prediction~\cite{tian2019fcos}.
For an $N$-way classification problem, the parameters $\gamma$ for the classifier normally produce $N+1$ logits for a softmax, corresponding to the $N$ classes and the background.
Meanwhile, the bounding box regressor's parameters $\beta$ contains $N$ stacked weights $\beta_c$, with one for each class $c$; the class with the highest prediction score determines which regressor's prediction is selected.
In order for our object detector to support the challenging iFSD setting, we make several key design choices affecting the two primary outputs of a detector: classification and localization.

\minisection{Incremental Classification Without Forgetting}
A major contributor to catastrophic forgetting is highly non-\textit{i.i.d.} sequential training with a shared classification head~\cite{farquhar2018towards}; optimizing the softmax can result in destructive gradients overwriting previous knowledge.
We thus replace the single softmax-based classifier $\mathcal{C}_{\gamma}$ with many binary sigmoid-based classifiers $\mathcal{C}_{\gamma_c}$, with each class individually handled by its own set of parameters. When trained with the focal loss~\cite{lin2017focal}, sigmoid classifiers have been shown to be just as effective as a single softmax classifier. 
Thus, when adding novel classes, we can train or generate a new set of classifier parameters ${\gamma_{c}^n}$.
When combined with previous parameters to predict all available classes, there is zero interference between each class's prediction score. 

\minisection{Class-agnostic Bounding Box Regressor}
Previous few-shot object detection methods~\cite{kang2019few, wang2020frustratingly, perez2020incremental} have tended to learn a per-class box regressor $\mathcal{B}_{\beta_c}$ in tandem with the classifier.
However, when only a few examples are available for learning, the model has very little opportunity to learn a custom location regressor for each novel class. 
Instead, we propose pretraining the base object detector with a single class-agnostic box regressor $\mathcal B_{\beta}$ for all classes.
When adapting the model to novel classes $C^n$, we simply reuse $\mathcal B_{\beta}$ for localization.
Such an approach has been shown to work well for zero-shot object detection if pretrained on a large-scale dataset~\cite{gu2021zero} and can leverage progress in the open-world detection literature~\cite{kim2021learning}. 
By alleviating the need to learn localization in a few-shot or continual manner, we can treat the problem as a few-shot classification task and focus just on generating additional classifier parameters ${\gamma_{c}^n}$.
We validate the effectiveness of this setup in Section~\ref{ablations_and_discussion}.

We can satisfy both the aforementioned objectives with FCOS~\cite{tian2019fcos}, a simple one-stage and anchor-free object detector. 
With these design choices, we decouple the few-shot novel class detection problem into serial tasks of localization and few-shot classification, dramatically simplifying it.

\vspace{-2mm}
\subsubsection{Few-shot Hypernetwork}
\label{few-shot hypernetwork}
\vspace{-2mm}
With localization handled by the class-agnostic object detector, the problem reduces to few-shot classification.
Sylph uses a hypernetwork $\mathcal{H}_\psi$ to generate parameters $\gamma^*_c = \{w_c, b_c\}$ for each binary classifier $\mathcal{C}_{\gamma^*_c}$.  
$\mathcal{H}_\psi$ takes as input an $N$-way $K$-shot episode of support set samples, consisting of $K$ instances of $N$ classes randomly sampled from the meta-training set.
We denote this support set $S^{N\times K} = (I^{N\times K}, b^{N\times K})$, with $I^{N\times K} \in \mathbb R^{(N* K)\times C \times H\times W}$.
The hypernetwork is modularized into three components: support set feature extraction, code prediction, and code aggregation and normalization, which we detail below.

\minisection{Support Set Feature Extraction}  The first stage consists of extracting features from the episode's support set. 
We share the same convolutional backbone $\mathcal{F}_\theta$ from the base detector to obtain features for each of the support set images, as it can be pretrained with the base detector in normal batch training. 
ROIAlignV2~\cite{he2017mask} is then used to pull the features corresponding to the location of each instance of each class.
We choose to crop at the feature level rather than at the image level, as features have a larger receptive field, potentially allowing for increased global context.
ROIAlignV2 produces a fixed size feature $z_{c,i} \in \mathbb{R}^{d_f \times d_h \times d_w}$ for each object instance, with $d_f$ being the channel dimension of the final layer of the backbone, and typically $d_h = d_w = 7$.
\begin{figure}
    \centering
    \includegraphics[width=0.75\columnwidth]{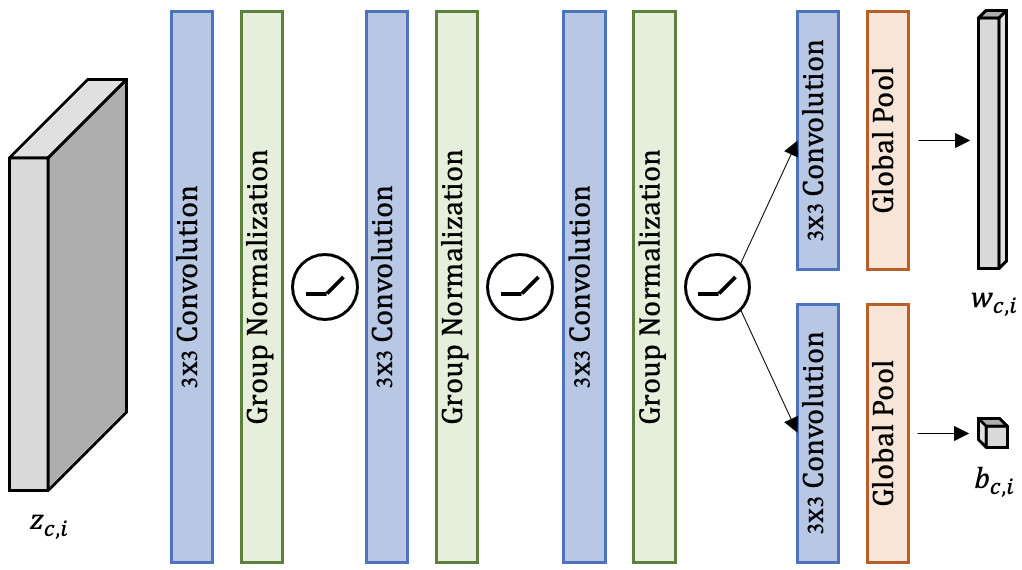}
    \caption{Hypernetwork architecture of the code predictor head. }
    \label{fig:cph}
\end{figure}

\minisection{Code Predictor Head (CPH)} Given each support sample's extracted features $z_{c,i}$, hypernetwork $\mathcal{H}_\psi$ predicts weights $w_{c,i}\in \mathbb{R}^{1\times C\times k\times k}$ and bias $b_{c,i}\in \mathbb{R}$, where $C$ is the preceding channel dimension and $k$ is the convolutional kernel size. 
The code predictor head (Figure~\ref{fig:cph}) consists of a shared subnetwork consisting of $3\times 3$ convolutional layers interleaved with group normalization~\cite{wu2018group} and ReLU activation functions, followed by a layer for predicting a weight and bias.
Global average pooling after the weight and bias predictor layers is used to reduce the predicted weights to the final dimensions. 
While the hypernetwork is capable of predicting weights $w_{c,i}$ of arbitrary size, we choose a kernel size of $k=1$ so that the generated weights can be used as either convolutional or linear layer weights, allowing compatibility with both region-based and dense detection.

\minisection{Code Process Module (CPM) }
In the CPM, we aggregate the predicted parameters for all samples of a class from CPH into a single set of weights $w_c$ and bias $b_c$.
We found a simple average of the weights and the bias across shots to be effective: $w_c = \frac{1}{K}\sum_{i=0}^{k-1}(w_{c,i})$ and $b_{c} = \frac{1}{K}\sum_{i=0}^{k-1}(b_{c,i})$.
However, directly using the class code $w_c$ in this form can cause gradient exploding, especially when stacking multiple convolutional layers between the input features and the final predictor head~\cite{gidaris2018dynamic}. As shown in Fig.~\ref{fig:loss_change}, gradient clipping~\cite{pascanu2013difficulty} can help, but occasionally the model still does not converge well, leading to high variance in model accuracy. 

Our weights $w_c$, as generated from the input support set features at this stage, more closely resemble feature maps than classifier weights. To this end, we want to avoid directly passing $w_{c}$ to the conditional classifier. Inspired by the success of $L^2$-normalized feature embeddings in parameterizing classifiers for zero-shot object detection~\cite{bansal2018zero, gu2021zero}, we explore incorporating $L^2$-normalization of the weights $\frac{w_{c}}{{||w_c||}}$. We normalize along the channel axis (in contrast to batch normalization~\cite{ioffe2015batch}) to ensure
weights for different classes do not interact. With normalization, learning is simplified and training is stabilized by mapping the support set features onto a unit sphere. 

To ensure compatibility of the normalized weights $\frac{w_c}{{||w_c||}}$ with a non-cosine classifier, we follow~\cite{salimans2016weight} and add a learnable scalar parameter $g$, rescaling the normalized weights as $w_c^*=\frac{g}{||w_c||}w_c$.
This allows us to avoid needing to adapt the classifier in the base detector.
By replacing per-class norm with a universal $g$, we end up with less variance across all class weights. We found that predicting the bias counteracts this negative effect. For the bias, we further add a prior bias $b_p = -\log((1-\pi)/\pi), \pi=0.01$ following~\cite{lin2017focal} and with a scalar $g_b$, resulting in a final bias of $b_c^* = g_b*b_c + b_p$.

\subsection{Training and Evaluation Details}
\vspace{-2mm}
To train the base detector and the hypernetwork, Sylph framework requires two sequential training stages: pretraining the base detector and learning the hypernetwork. 

\minisection{Base Object Detector Pretraining}
We first pretrain the base detector $D_{\phi}$ with batch stochastic gradient descent on base classes $C^b$, optimizing for classification and bounding box regression losses. 
We choose FCOS as our base detector; we refer the reader to \cite{tian2019fcos} for further training details. 
The pretraining process produces trained parameters $\theta$ and $\phi$, as well as class agnostic box regression parameters $\beta$ and class codes for the base classes $\gamma_b = \{w_{c_b}, b_{c_b}\} ~ \forall ~ {c_b} \in C^b$.
Thus, at the conclusion of this stage, we have a detector $D_\phi$ capable of producing bounding boxes in an image for the base classes and potentially novel classes as well.

\minisection{Meta-training} 
During meta-training, we create few-shot episodes of $N$ categories by sampling a set of $N\times (K+1)$ image and bounding box tuples $(I, b)$ from $C^b$, a support set of $N \times K$ samples, and a query set with $N \times 1$ samples. The query set is used as input to the base detector. Only the focal loss~\cite{lin2017focal} from the classification branch is computed at this stage.
The primary goal at this stage is to train the few-shot hypernetwork $\mathcal{H}_\psi$ so that it is able to map $S^{N\times K}$ to a new set of synthesized class codes $\gamma^*_{c_b} = (w^*_{c_b},b^*_{c_b})$ for classification. 
We freeze the whole base object detector except the four convolutional layers in the FCOS classification subnetwork and replace its initial classifier with our conditional classifier capable of taking the synthesized class codes to make predictions on the query image features. 
We found that finetuning these extra convolutional layers in the base detector results in better overall performance than not finetuning them (Section~\ref{ablations_and_discussion}).
In preliminary experiments we found
that the more components/layers we initialize from pretraining the better our final AP for base classes.

\minisection{Meta-testing} 
To evaluate the model's performance across all classes, we take $K$ shots per-class samples from the whole set and make feed-forward passes through the hypernetwork one class at a time to synthesize class codes $\gamma^*_c = \{w^*_{c}, b^*_{c}\} ~ \forall ~ {c} \in C^b \cup C^n $. 
With the generated codes, the base detector is able to perform inference at the same inference speed and behavior as a normal detector. This setup of our model is denoted as \textit{Sylph}.

\section{Experiments}
\label{headings}
\vspace{-2mm} 

\begin{table*}[t]
\begin{center}
\caption{Benchmarking on the eval split of LVIS-v1.
~We use $K=10$ shots to infer base class codes and all available data for the rare classes ($\leq 10$). Both \textit{ONCE}$^*$ and \textit{Sylph} predict all classes in a single pass. The base and novel data checkmarks indicate whether the data is used to update model weights during an incremental learning step. 
}
\label{table:lvis_meta_train}
\vspace{-2mm}
\resizebox{0.93\textwidth}{!}{%
\begin{tabular}{c|lccccllll}
\multicolumn{1}{c|}{\bf Pretrain} &  \multicolumn{1}{c}{\bf Method}& \multicolumn{1}{c}{\bf Base Data} & \multicolumn{1}{c}{\bf Novel Data} &
\multicolumn{1}{c}{\bf Continuous} & \multicolumn{1}{c}{\bf Re-training}
&  \multicolumn{1}{c}{$\mathbf{AP}$}&\multicolumn{1}{c}{$\mathbf{AP_r}$} &
\multicolumn{1}{c}{$\mathbf{AP_c}$} &
\multicolumn{1}{c}{$\mathbf{AP_f}$}  
\\ 
\toprule

\multirow{4}{*}{Default}
 & ONCE$^*$~\cite{perez2020incremental} & & & \checkmark    &    & 12.9 ($\pm$0.65)& 6.3 ($\pm$0.38) & 11.2 ($\pm$0.60) &17.7 ($\pm$0.97) \\ 

& \bf Sylph &    & & \checkmark & &   \bf 18.5 ($\pm$0.12)(\textcolor{green}{$\uparrow$}5.6) & \bf 10.0 ($\pm$0.17) & \bf 16.5 ($\pm$0.25)  & \bf 24.3 ($\pm$0.12) \\
\cmidrule(lr){2-10}

 & TFA-ours & &\checkmark  &      &\checkmark & 21.0 & \bf11.9 &17.7 & 28.6\\ 
& TFA$^*$~\cite{wang2020frustratingly} & \checkmark($K$ shots) & \checkmark & & \checkmark & \bf 21.1 & 9.1 & \bf21.6 & 25.9 \\

& Joint-train~\cite{gupta2019lvis} &\checkmark & \checkmark & &\checkmark  & 20.7 & 8.7 &18.2 & \bf 28.8\\ 
\hline

\multirow{4}{*}{Aug}& ONCE$^*$~\cite{perez2020incremental} & & &\checkmark         & & 8.5 ($\pm$0.24)& 6.1 ($\pm$0.31) & 7.8 ($\pm$0.42) &10.2 ($\pm$0.14)\\ 

& \bf Sylph & & &\checkmark& &   \bf 20.7 ($\pm0.10$)(\textcolor{green}{$\uparrow$}12.2)&\bf 13.9 ($\pm0.21$) &\bf 19.0 ($\pm0.19$) & \bf 25.5 ($\pm0.02$) \\
\cmidrule(lr){2-10}
& TFA-ours-aug & &\checkmark &  &\checkmark & \bf 25.1 &\bf 17.8 &22.6&\bf 30.9  \\ 
& TFA$^*$-aug~\cite{wang2020frustratingly} &\checkmark($K$ shots) & \checkmark & & \checkmark& 24.4 & 16.1 & \bf 24.9 & 27.6 \\

& Joint-train~\cite{gupta2019lvis} &\checkmark & \checkmark &  &\checkmark& 24.3 &13.3&22.7 & \bf 30.9\\

\hline

\multirow{4}{*}{All} & ONCE$^*$~\cite{perez2020incremental}  & &  & \checkmark      & & 19.4 ($\pm0.08)$ & 12.3 ($\pm0.33)$ &18.8 ($\pm0.28)$ & 23.3 ($\pm0.12)$\\ 
 & \bf Sylph & & & \checkmark & & 
 \bf 24.6 ($\pm0.10$)(\textcolor{green}{$\uparrow$}
 5.2)&\bf 16.5 ($\pm0.34$) &\bf 23.7 ($\pm0.17$) & \bf 29.1 ($\pm0.02$) \\ 
 \cmidrule(lr){2-10}

& TFA-ours-aug & & \checkmark &  & \checkmark& \bf 27.5 &19.3 &25.6 &33.0  \\ 
& TFA$^*$-aug~\cite{wang2020frustratingly} &\checkmark($K$ shots) & \checkmark & & \checkmark& 27.0 &\bf 19.7 &\bf 27.6 &29.6 \\ 

& Joint-train~\cite{gupta2019lvis} &\checkmark & \checkmark  & & \checkmark &27.2 & 18.0 & 26.4& \bf 33.6 \\
\bottomrule
\end{tabular}
}
\vspace{-4mm}
\end{center}
\end{table*}

\minisection{Datasets and Metrics} We benchmark and ablate Sylph on two datasets: COCO~\cite{lin2014microsoft} and LVIS~\cite{gupta2019lvis}. 
For \textbf{COCO}, we follow the split commonly used for few-shot object detection~\cite{kang2019few,perez2020incremental,wang2020frustratingly}: the 60 categories that are disjoint from PASCAL VOC~\cite{everingham2010pascal} are used as base classes, while the remaining 20 classes are designated novel. We report experimental results for $K = \{1, 5, 10, 20, 30\}$ shots on the COCO minival set. 
For \textbf{LVIS-v1}, we follow the organically long-tail distribution of the dataset as proposed in~\cite{wang2020frustratingly} to produce a  base-novel split. LVIS contains 405 frequent classes appearing in more than 100 images, 461 common classes with 10-100 images, and 337 rare classes with fewer than 10 images, for a total of 1203 object categories. In our experiments, we use the 337 rare classes as novel classes and the 866 frequent and common classes as base classes. 

For evaluation metrics, we report mean average precision (mAP) computed on a per-split basis; we run inference for both the base and novel classes in a single pass, but we report mAP separately as different models tend to have different performances across splits. 
For COCO, we denote the mAPs for base and novel categories as $AP_b$ and $AP_n$, respectively. Similarly, for LVIS, $AP_r$, $AP_c$, and $AP_f$ is the average precision aggregated across rare, common, and frequent classes, respectively. For all experiments, we report the mean and standard deviation of the mAP across five meta-testing runs.
We run experiments with several pretraining strategies: (1) \textit{Default}: the model is pretrained on ImageNet-1k~\cite{russakovsky2015imagenet}; (2) \textit{Aug}: large-scale jittering (LSJ)~\cite{ghiasi2021simple} and RandAugment~\cite{cubuk2020randaugment} are also applied; and (3) \textit{All}: in addition to the aforementioned augmentations, IG-50M pretrained backbone weights from PreDet~\cite{ramanathan2021predet} are used.

\minisection{Implementation Details}
For all our experiments, we use a ResNet-50 backbone \cite{he2016deep} with a feature pyramid network (FPN)~\cite{lin2017feature}. We use SGD with momentum ($0.9$) and weight decay ($1\mathrm{e}{-4}$) for all training stages. During pretraining, we set the learning rate to $1\mathrm{e}{-2}$ with a batch size of 16; we increase the batch size to 128 when data augmentation is on. During meta-training, we set the learning rate ($lr$) to $5\mathrm{e}{-4}$. We uniformly sample 3-way 5-shot tasks from the base classes, with a single query image per class.

We pretrain for 90k steps ($\sim$11hrs), with an extra 30k steps for meta-learning ($\sim$13hrs). The $lr$ is decreased tenfold at steps 60k and 80k in the pretraining, and at 20k and 26k during the meta-training. Finally, we limit the number of detections per image to 100 for COCO and 300 for LVIS.
We build our framework on top of Detectron2~\cite{wu2019detectron2}; we plan to publicly release our code upon publication.

\begin{table}[t]
\caption{Benchmarking on COCO Dataset, evaluated on minival set.
We benchmark Sylph against ONCE for $K={1,5,10}$ shots, with additional $K={20, 30}$ shots for Sylph.
We also include 10-shot TFA, which finetunes on novel data. To mimic the training protocol of ONCE, we apply early-stop (at 30k steps) to Sylph pretraining (denoted Sylph-es).}
\label{coco-meta-train}
\vspace{-4mm}
\begin{center}
\resizebox{0.8\columnwidth}{!}
{%
\begin{tabular}{c|lll}
\multicolumn{1}{c|}{\bf Shot} & \textbf{Method}  &\multicolumn{1}{c}{$\mathbf{AP_n}$} &\multicolumn{1}{c}{$\mathbf{AP_b}$}
\\ \toprule
\multirow{3}{*}{1} 
&ONCE~\cite{perez2020incremental}             &0.7 & 17.9 \\
&\textbf{Sylph}    &\bf 0.9 ($\pm$0.11) & \bf 29.8 ($\pm$1.17) \\  
&\textbf{Sylph (All)}  &\bf 1.1 ($\pm$0.14) & \bf 37.6 ($\pm$1.57) \\   
\hline
\multirow{3}{*}{5} 
&ONCE~\cite{perez2020incremental}             &1.0 &17.9 \\
&\textbf{Sylph}    & 1.4 ($\pm$0.12)  & 35.5 ($\pm$0.18) \\   
&\textbf{Sylph (All)}    &\bf 1.5 ($\pm$0.05) & \bf 42.4 ($\pm$0.13) \\  
\hline       
\multirow{7}{*}{10} 
&ONCE ~\cite{perez2020incremental}            &1.2 &17.9 \\
&\textbf{Sylph}    &1.6 ($\pm$0.06) &35.8 ($\pm$0.05) \\   
&\textbf{Sylph-es}   &2.3 &22.4\\
&\textbf{Sylph (All)}    &1.7 ($\pm$0.05) & \bf 42.8 ($\pm$0.07) \\  
&\textbf{Sylph-LVIS}    &\bf 3.8 ($\pm$0.20) & 37.7\\  
\cmidrule(lr){2-4}
&Joint-train~\cite{gupta2019lvis}      &4.0 & 37.7 \\
&TFA-ours      & 3.6 & N/A \\
&TFA$^{*}$~\cite{wang2020frustratingly}      & 5.7 & 35.9 \\ 

\hline
\multirow{1}{*}{20} 
&\textbf{Sylph}    &1.62 ($\pm$0.06) &36.0 ($\pm$0.08) \\   
                   
\hline
\multirow{1}{*}{30} 
&\textbf{Sylph}    &1.65 ($\pm$0.06) &36.1 ($\pm$0.08) \\   
\bottomrule
\end{tabular}
}
\vspace{-5mm}
\end{center}
\end{table}

\subsection{Incremental Few-shot Object Detection}
\label{experiment_benchmarking}
\vspace{-2mm}
As the only other method designed for iFSD, we primarily compare against ONCE on both COCO and LVIS.
Focusing on the \textbf{finetuning-free iFSD} evaluation protocol~\cite{perez2020incremental}, we demonstrate the effectiveness of Sylph with several pretraining strategies.
In addition, we report results for a few training-intensive FSD methods as an upper bound of our finetune-free approach, including joint-training, which is normally used for long-tailed datasets~\cite{gupta2019lvis}, and a finetuning-based method known as TFA~\cite{wang2020frustratingly}.

\minisection{Finetuning-free iFSD benchmarking}
We primarily compare with ONCE~\cite{perez2020incremental}, as the most relevant method in this setting.
On COCO, we compare with the reported number from~\cite{perez2020incremental} in Table~\ref{coco-meta-train}. For LVIS, we re-implement ONCE with a baseline version of our code generator which has no bias prediction, no weight norm, no scaling factor $g$ in the CPM, and no convolutional layers in the shared portion of the CPH, which effectively leaves the basic components of the hypernetwork as close to the originally-proposed ONCE as possible. 
We denote this version ONCE$^*$ in Table~\ref{table:lvis_meta_train}.

We demonstrate that the key design choices of Sylph allow it to significantly outperform ONCE on both datasets, across all data splits. On the large-scale dataset LVIS, \textit{Sylph} surpasses ONCE by 8\% averaged across different pretraining strategies in a fair head-to-head (no additional data augmentation or pretraining data). For the heavy data augmentation setup, \textit{Aug}, ONCE$^*$ struggles to converge during training, resulting in much worse performance than Sylph. 
In particular, we show that our method is truly able to learn novel categories from few shots without forgetting base classes. 
For example, with early stopping during pretraining (Sylph-es in Table~\ref{coco-meta-train}) and $K=10$ shots on COCO, we achieve an $AP_n$ twice as good as ONCE, while still surpassing it by 4 points for the base classes. 

\minisection{Joint-training and finetune-based iFSD as upper bounds} 
For the \textit{Joint-train} method, we follow~\cite{gupta2019lvis} to ensure its effectiveness on the novel split in the low-data regime. In particular, we perform repeat factor sampling with the factor set to 0.001 in order to balance the sampling frequency across different classes during training.
We select TFA~\cite{wang2020frustratingly} to represent finetuning-based iFSD methods. 
For this, we adapt the TFA~\cite{wang2020frustratingly} methodology to our FCOS-based framework, following the finetuning protocol as closely as possible. Specifically, this involves two training steps: (1) pretraining of the base detector on base classes; (2) sampling $K=10$ shots across both base and novel classes while freezing all layers other than the box regressor and the classifier. We finetune the regressor and train a new classifier for all classes, with base classifier parameters initialized from pretraining. We denote the adapted TFA method as \textit{TFA$^*$}. Additionally, we make several modification to the standard TFA to bring it closer to our setup, adjusting to an incremental batch setup~\cite{perez2020incremental} where the novel classes are added in a single round.  In particular, in the finetuning stage, (1) only $C^n$ is used, (2) the box regressor is kept frozen, and (3) the classifier is not initialized with any pretrained base class parameters, as we do not finetune on the base classes. In this setup, a deployed model can be directly extended without any backbone modification to novel categories. We label this version of TFA as \textit{TFA-ours}, as it is made possible by our framework. 
As our model relies on a large-scale dataset in the meta-training stage, we create a variant, \textit{Sylph-LVIS}, which uses the LVIS dataset excluding the COCO novel classes while keeping all parts of the base detector frozen so that it is able to preserve its pretrained COCO base class codes. 

We report these benchmarking results on both datasets in Tables~\ref{table:lvis_meta_train} and \ref{coco-meta-train}. An interesting observation here is that, for all the FSD methods we benchmarked, their precision on the novel split increases significantly with an increased number of base classes. As we go from the 60 base classes in COCO to 866 base classes in LVIS, \textit{Sylph} achieves around a 9\% gain in $AP_r$ and $AP_n$ in Tables~\ref{table:lvis_meta_train} and \ref{coco-meta-train}, even surpassing the gains achieved by all the variants where training and finetuning is allowed. 
In terms of the overall precision across all classes, \textit{Sylph}'s performance is not far from finetuning-intensive methods: just 3 and 4 points lower compared to the best performing upper bound method on LVIS. On COCO, for example, with class augmentation in the meta-learning stage, our \textit{Sylph-LVIS} achieves 3.8 AP, only 0.2 AP short of the joint-training approach. 

\minisection{More FSD model behavior analysis} On LVIS, we benchmarked all FSD methods across three pretraining strategies, showing that all methods benefit from the use of additional augmentations and large-scale weak supervision pretraining across both novel and base classes, including \textit{Sylph}. This behavior is highly desirable for \textit{Sylph}, as it benefits from any improvement to the base detector.

However, for a smaller scale dataset (\eg COCO), the boost in novel class performance for \textit{Sylph} is less than the gain on the base classes, as shown in Table~\ref{coco-meta-train} for rows corresponding to \textit{Sylph (All)}. This is related to episodic learning, where more tasks lead to improved learning compared to more per-task data. We further validate this with \textit{Sylph-LVIS}, which has a similar amount of training data, but with more tasks; impressively, we find that \textit{Sylph-LVIS} achieves comparable accuracy to the joint-training approach. 
Still, we see a large precision gap between \textit{Sylph-LVIS} on the COCO novel split and \textit{Sylph} on LVIS rare classes, indicating that large-scale pretraining is essential, as it results in (1) a more accurate bounding box locator, and (2) a feature extractor that can generalize better to novel classes. 

Also, surprisingly, the adapted simple approach \textit{TFA-ours} is able to achieve better precision on LVIS than its standard counterpart \textit{TFA} with our selected base detector, with the advantage of not revisiting the base data at all.

\vspace{-2mm}
\section{Ablations and Further Discussion}
\label{ablations_and_discussion}
\vspace{-2mm}
We run all experiments in this section with the \textit{Default} pretraining strategy unless otherwise stated.

\begin{figure}[t!]
    \centering
    \includegraphics[width=0.9\columnwidth]{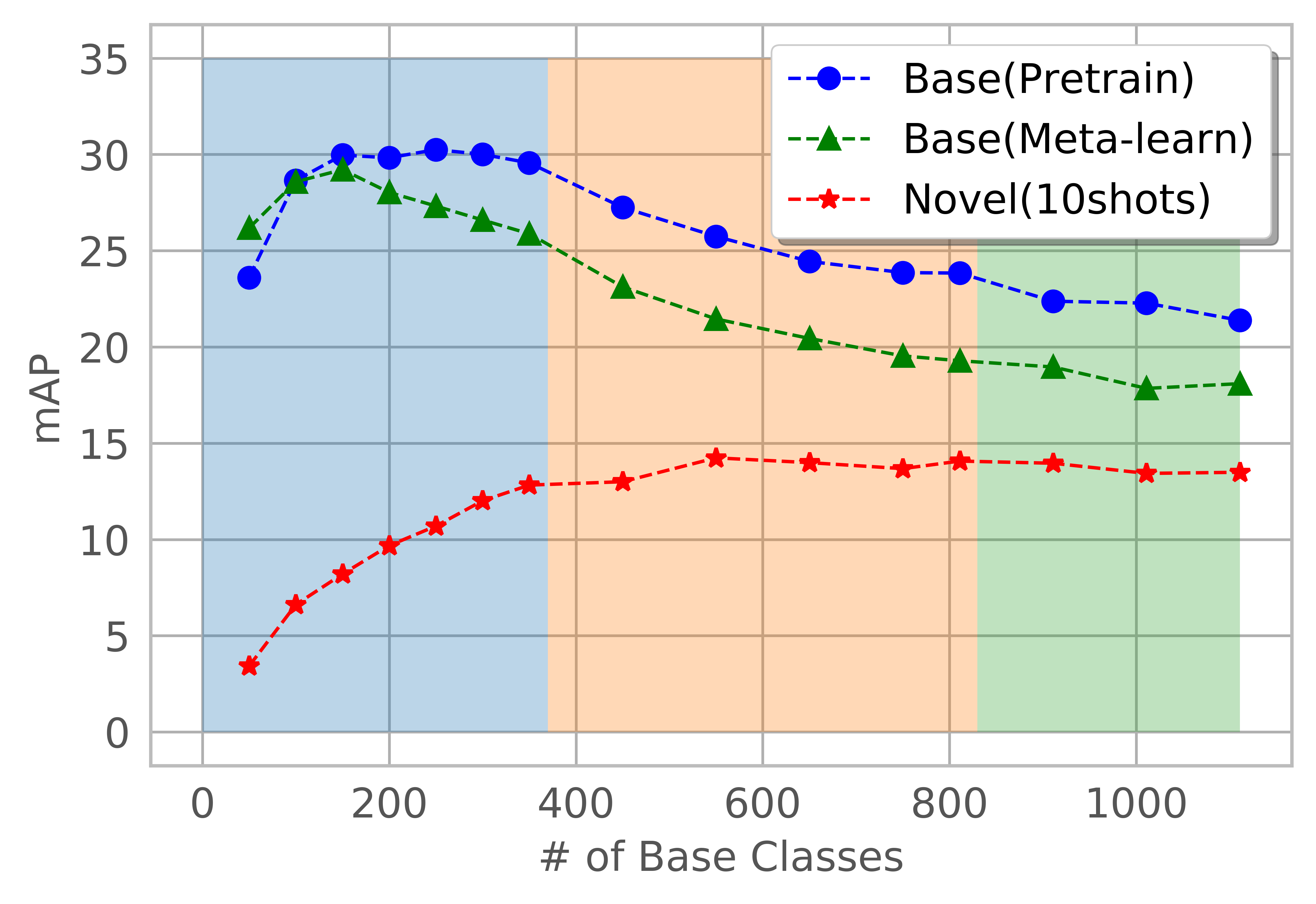}
    \caption{ The effect of the number of base classes in \textit{sylph} meta-training. The blue, orange, and green backgrounds denote frequent, common, and rare classes, respectively. }
    \label{fig:num_base_classes_impact}
    \vspace{-4mm}
\end{figure}

\minisection{How does the number of base classes impact the novel class precision?}  
For this experiment, we randomly choose 50 classes from the frequent classes of LVIS as a novel set. We report base and novel mAP in Figure~\ref{fig:num_base_classes_impact} on this fixed novel while gradually increasing the number of base classes, starting from the frequent, moving to common, and then the rare classes. For all the plotted points, we complete the training of \textit{Sylph} on the base split, and use $K=10$ for class codes inference.
We confirm the effect of a larger base set in novel class detection precision in Fig.~\ref{fig:num_base_classes_impact}. Indeed, novel class mAP rapidly increases in the frequent classes region, slows down when adding common classes, and starts to fully stabilize at around 800 base classes in the rare classes region. 
Not surprisingly, novel class score increases more rapidly when the backbone is pretrained with classes that have more samples.
These results indicate for the first time that challenging incremental few-shot detection is feasible when there is a large enough base dataset.

\begin{figure}[t]
    \centering
    \includegraphics[width=0.46
    \textwidth]{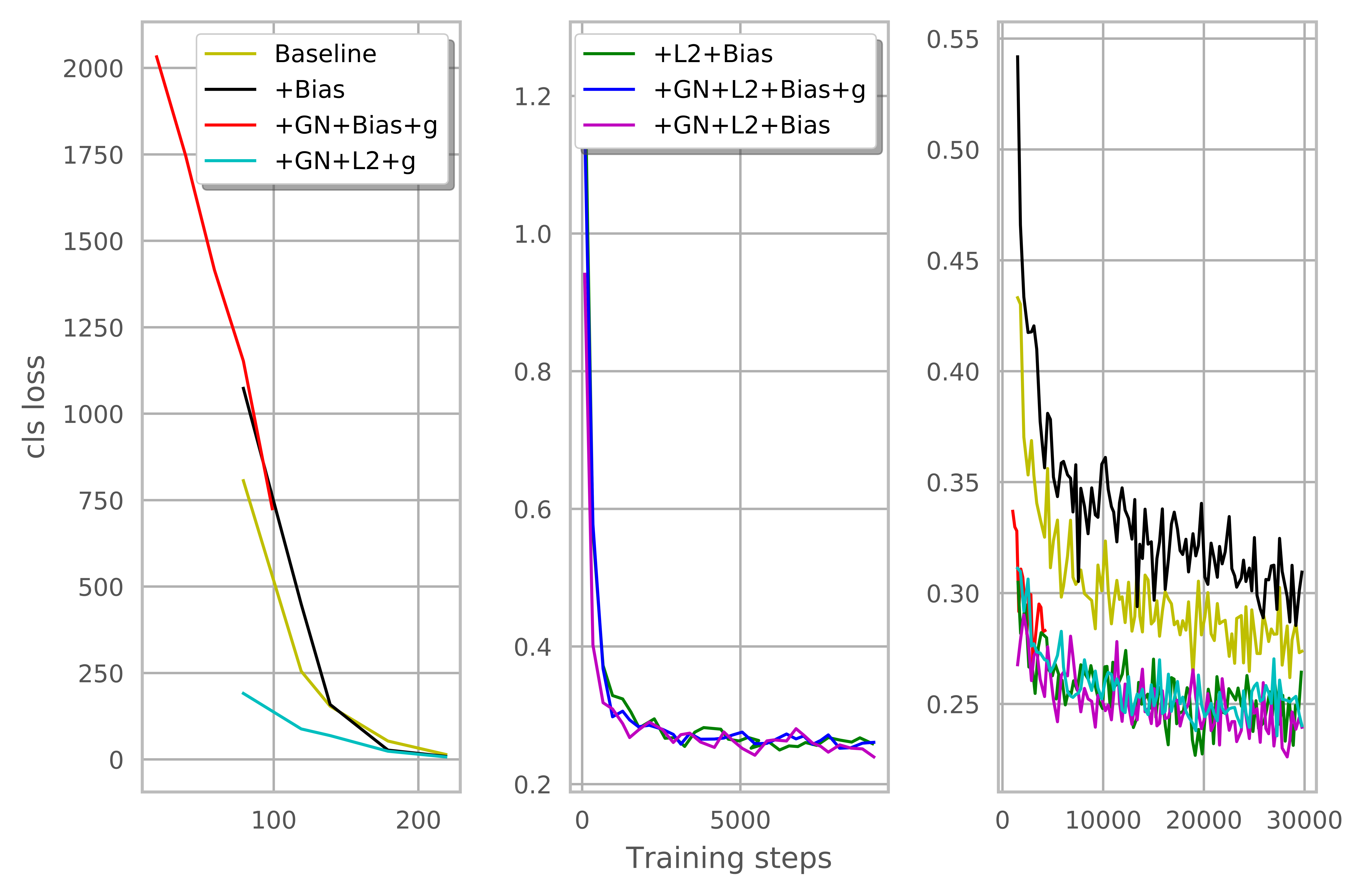}
    \vspace{-2mm}
    \caption{The loss comparison for different model setups. 
    On the right side, we plot the losses starting from training step 100.
    }
    \label{fig:loss_change}
    \vspace{-1mm}
\end{figure}

\minisection{Model ablations} We study the effect of several key elements of our model, including the normalization scheme (GroupNorm (GN) vs $L^2$-Norm), the weight scaling factor $g$,  predicted bias, and the number of convolutional layers stacked in CPH for the classifier. 
These results are shown in Table~\ref{ablation_modeling}, from which we can see that overall, using either $L^2$-Norm or GN is very beneficial for Sylph, improving the overall AP across all subsets of classes by around 6 absolute points w.r.t. the baseline. However, when both GN and $L^2$-Norm are applied, there is no obvious extra gain. Comparing the last three rows in Table~\ref{ablation_modeling}, we see that the use of $g$ and bias results in a small accuracy improvement.  We plotted the loss of different configurations in Fig.~\ref{fig:loss_change}. 
From the leftmost figure, we can see that $L^2$-Norm has the largest impact in curbing the loss than any other configuration. Additionally, from the rightmost figure, we can see that both $L^2$-Norm and GN converge better than models without normalization.
Overall, we can conclude that the elements that form \textit{Sylph} are effective at both learning base classes and generalizing to novel ones.

\begin{table}[!t]
\caption{Ablation study: Modeling choices of \textit{Sylph} on the LVIS.  }
\vspace{-4mm}
\begin{center}
\label{ablation_modeling}
\resizebox{1\columnwidth}{!}{%
\begin{tabular}{ccccccccc}
\multicolumn{1}{c}{\bf Bias} & \multicolumn{1}{c}{\bf GN} & \multicolumn{1}{c}{\bf{$L^2$-Norm}} & \multicolumn{1}{c}{$g$}& \multicolumn{1}{c}{\bf $\#$conv} &\multicolumn{1}{c}{\bf AP}  &\multicolumn{1}{c}{$\mathbf{AP_r}$}&\multicolumn{1}{c}{$\mathbf{AP_c}$} &\multicolumn{1}{c}{$\mathbf{AP_f}$}
\\ \toprule
  &  &     &  & 0 & 12.9 ($\pm$0.65)& 6.3 ($\pm$0.38) & 11.2 ($\pm$0.60) &17.7 ($\pm$0.97) \\ 
 \checkmark &  &    &  & 0 & 9.7 ($\pm$0.25)& 3.9 ($\pm$0.37) & 8.3 ($\pm$0.27) &13.7 ($\pm$0.34) \\ 
\checkmark& \checkmark & & \checkmark &0 & 17.2 ($\pm$0.12)  & 8.8 ($\pm$0.26) & 15.3 ($\pm$0.23) &  23.2 ($\pm$0.14)\\
\checkmark&  & \checkmark & \checkmark& 0 & 18.0 ($\pm$0.12)  & 9.3 ($\pm$0.39) & 16.1 ($\pm$0.15) &  24.0 ($\pm$0.15)\\ 
\checkmark& \checkmark  & \checkmark & \checkmark &0 & 18.0 ($\pm$0.12)  & 9.1 ($\pm$0.29) & 16.2 ($\pm$0.27) &  24.0 ($\pm$0.15)\\ 
 \checkmark & \checkmark  & \checkmark &\checkmark& 1 & 18.5 ($\pm$0.12)  & 10.0 ($\pm$0.17) & 16.5 ($\pm$0.25) &  24.3 ($\pm$0.12)\\
 \checkmark & \checkmark  & \checkmark & \checkmark& 2 &  \bf 18.6 ($\pm$0.13) & \bf 10.1 ($\pm$0.16) &  16.7 ($\pm$0.29)  &  \bf \bf 24.5 ($\pm$0.10) \\ 
 \checkmark & \checkmark  & \checkmark & & 2 & 18.5 ($\pm$0.07) &  9.0 ($\pm$0.17) & \bf 16.8 ($\pm$0.20)  & \bf 24.5 ($\pm$0.08) \\
 & \checkmark  & \checkmark & & 2 &  18.4 ($\pm$0.07) &  9.5 ($\pm$0.29) &  16.7 ($\pm$0.17)  & 24.1 ($\pm$0.11) \\ 
\bottomrule
\end{tabular}
}
\vspace{-3mm}
\end{center}
\end{table}

\minisection{Training recipe ablations} We also explored different training recipes, (1) \textit{FA}: strictly freezing the whole base detector during meta-training, preserving the pretrained base class codes. (2) \textit{Joint}: pretraining and meta-training on all available classes with the default setup. We report the result in Table~\ref{ablation-training-recipe}. As we can see, \textit{Sylph}'s training recipe beats that of \textit{FA} by 2 points on the \textit{All} setup. This means that allowing the classification convolutional subnetwork in the base detector to adapt during meta-training is important in our proposed framework. However, \textit{Joint} performs comparably to $FA$, falling behind \textit{Sylph} even though it has seen the novel classes during training. We think this might be  explained by two reasons: (1) As the number of base classes increases, \textit{Joint} struggles to recover the base class AP in meta-training. (2) As we use uniform sampling on the class-level, when mixed with rare classes, the more frequent classes get sampled less, thus leading to AP drop on those splits.

\minisection{Does freezing the base detector in the meta-test stage limit few-shot continual learning capabilities?} We set up a simple two step continual learning task and solve it with finetuning~\cite{parisi2019continual, wang2020frustratingly}. In particular, given a pretrained FCOS model on base classes, we freeze most parts of the detector and finetune the remaining parts on all available novel data.  
On COCO, we keep the same base and novel splits. On LVIS, we use 100 randomly selected frequent classes as the novel split and the remaining 1103 classes as base classes. We follow \textit{TFA$^*$-st} where the box regressor and  classifier are finetuned and \textit{TFA-ours} where only the classifier is trained. 
The results, along with a normal training on the novel set from scratch, are shown in Table~\ref{table:pre-train-finetune-novel}. We see that, surprisingly, there is no obvious performance drop for the finetune approach, and even with the strict setup \textit{TFA-ours},  the $AP_n$ only decreases by 3 points. As \textit{TFA-ours} closely resembles the training scheme of \textit{Sylph}, we can conclude that the formulation we propose here does not entail a large sacrifice to the novel class learning potential.

\begin{table}[t!]
\caption{Effect of the training recipe on the Sylph Framework. We report average precision across five runs.}
\vspace{-4mm}
\begin{center}
\label{ablation-training-recipe}
\resizebox{0.7\columnwidth}{!}{%
\begin{tabular}{c|clcccc} \multicolumn{1}{c|}{\bf{Setup}}   & \bf Method &\multicolumn{1}{c}{\bf AP}  &\multicolumn{1}{c}{$\mathbf{AP_r}$}&\multicolumn{1}{c}{$\mathbf{AP_c}$} &\multicolumn{1}{c}{$\mathbf{AP_f}$}
\\ \toprule
\multirow{3}{*}{Default}  & FA & 18.3  & 10.5 & 16.3 &  23.9 \\
& Joint  &  \bf 18.5 & \bf 10.8 & \bf 16.5  & 24.2 \\ 
& Sylph  &  \bf 18.5 & 10.0 & \bf 16.5  & \bf 24.3 \\ 

 \hline
 
 \multirow{2}{*}{Aug} & FA& 19.9&12.5 &18.5& 24.8 \\
 & Joint   &  19.2 & 12.8 & 17.7  & 23.8 \\ 
 & Sylph & \bf 20.7&\bf 13.9 &\bf 19.0 & \bf 25.5 \\
 \hline
 \multirow{2}{*}{All} &FA & 22.5&15.4 &21.2 & 27.1 \\ 
 & Joint   & 22.5 & \bf 17.2 &  21.2  & 26.4\\ 
 & Sylph & \bf 24.6&16.5 &\bf23.7 & \bf 29.1 \\
\bottomrule
\end{tabular}
}
\vspace{-3mm}
\end{center}
\end{table}

\begin{table}[!t]
\caption{
Novel set accuracy comparison across the finetuning approaches. Full training of the detector is denoted \textit{Scratch}.
}
\vspace{-4mm}
\begin{center}
\label{table:pre-train-finetune-novel}
\resizebox{0.7\columnwidth}{!}{%
\begin{tabular}{c|c|cccc}
\multicolumn{1}{c|}{\bf Dataset} &\multicolumn{1}{c|}{\textbf{Setup}} & \textbf{Method} & \textbf{Cls} & \textbf{Box}  &\multicolumn{1}{c}{\bf APn} 
\\ \toprule
\multirow{3}{*}{COCO} &
 \multirow{3}{*}{All}  &Scratch & N/A & N/A & 49.9 \\
& &TFA-ours & \checkmark & & 47.1 \\
& &TFA$^*$-st & \checkmark & \checkmark & 50.5\\ \hline
\multirow{3}{*}{LVIS} 
&      \multirow{3}{*}{All}  &Scratch & N/A & N/A & 34.9 \\
& &TFA-ours & \checkmark & & 30.5 \\
& &TFA$^*$-st & \checkmark & \checkmark & 34.3\\
\bottomrule
\end{tabular}
}
\vspace{-5mm}
\end{center}
\end{table}

\section{Conclusion}
We introduce Sylph, an object detection framework capable of extending to new classes from only a few examples in a continual manner without any training.
We empirically validate that our design choices lead to effective training and improved accuracy, showing for the first time that an iFSD without test-time training can achieve performance close to finetune-based methods on large scale datasets like LVIS.
While we view Sylph as an improvement over existing methods, there are limitations.
Though we have demonstrated that pretraining a class-agnostic detector can surface novel objects with high recall, it is not infallible and still dependent on large-scale datasets. 
Unlabeled objects due to annotator error or a class not being in the label set can result in false negatives in the dataset, which may lead to the model failing to surface such objects~\cite{yang2020object, joseph2021towards, konan2022extending}.
Additionally, more sophisticated aggregation methods to fuse support set features may also lead to further improvements.

\subsection*{Acknowledgements}
We thank Vignesh Ramanathan, Abhijit Ogale, and Zhicheng Yan for valuable discussions and insights.

%%%%%%%%% REFERENCES
{\small
\bibliographystyle{ieee_fullname}
\bibliography{bib}
}

\end{document}